\documentclass[preprint,12pt,authoryear]{elsarticle}

\usepackage{amssymb}

\journal{Computers and Electronics in Agriculture}

\usepackage{amsmath}

\usepackage{url}

\usepackage{xcolor}

\usepackage{enumitem}
\usepackage{natbib}

\usepackage{tabularx,booktabs}
\newcolumntype{Y}{>{\centering\arraybackslash}X}

\begin{document}

\begin{frontmatter}

\title{Extreme Gradient Boosting for Yield Estimation compared with Deep Learning Approaches}

\author[inst1,mail]{Florian Huber}

\affiliation[inst1]{organization={Department of Computer Science IV},
            addressline={University of Bonn}, 
            city={Friedrich-Hirzebruch-Allee 8},
            postcode={53121}, 
            state={Bonn},
            country={Germany}}

\author[inst1]{Artem Yushchenko}
\author[inst1,inst2]{Benedikt Stratmann}
\author[inst1]{Volker Steinhage}

\affiliation[inst2]{organization={Now with: University of Karlsruhe},
            postcode={76131}, 
            state={Karlsruhe},
            country={Germany}}
            
\affiliation[mail]{organization={Correspondence}, 
            country={huber@cs.uni-bonn.de}}

\begin{abstract}
Accurate prediction of crop yield before harvest is of great importance for crop logistics, market planning, and food distribution around the world. Yield prediction requires monitoring of phenological and climatic characteristics over extended time periods to model the complex relations involved in crop development. Remote sensing satellite images provided by various satellites circumnavigating the world are a cheap and reliable way to obtain data for yield prediction. 
The field of yield prediction is currently dominated by Deep Learning approaches. While the accuracies reached with those approaches are promising, the needed amounts of data and the ``black-box'' nature can restrict the application of Deep Learning methods. The limitations can be overcome by proposing a pipeline to process remote sensing images into feature-based representations that allow the employment of Extreme Gradient Boosting (XGBoost) for yield prediction. A comparative evaluation of soybean yield prediction within the United States shows promising prediction accuracies compared to state-of-the-art yield prediction systems based on Deep Learning. Feature importances expose the near-infrared spectrum of light as an important feature within our models. The reported results hint at the capabilities of XGBoost for yield prediction and encourage future experiments with XGBoost for yield prediction on other crops in regions all around the world.
\end{abstract}

\begin{keyword}
Yield Prediction \sep Remote Sensing \sep Extreme Gradient Boosting \sep Explainability \sep Shapley Value
\end{keyword}

\end{frontmatter}


\section{Introduction}

Accurate prediction of yields in agriculture comes with multiple benefits, such as advanced logistic planning capabilities, the possibility of redistributing food around the world, and financial security for producers. Obtaining reliable data for the task of yield prediction often goes along with monitoring phenological and climatic features over extended time periods and wide spatial spreading. Remote sensing satellite images provide such data in a cheap and reliable manner. Multispectral images give insight into the phenological states of plants, together with information about climatic conditions throughout the year, while being available worldwide. Machine Learning enables the employment of this information through its ability to efficiently process highly detailed and complex data.
\par
Deep Learning approaches dominate the field of yield prediction on remote sensing data so far (Section \ref{sec:deeplearningforyieldprediction}). Different architectures are used to process the complex relations provided by multispectral images and give respectable accuracies for yield predictions. These approaches work especially well, when data is plentiful and no extensive explanations of the results are needed, due to the ``black-box'' nature of Deep Learning.
\par
This study aims to reproduce the recent success of Extreme Gradient Boosting (XGBoost) \citep{chen2016xgboost} in the field of yield prediction. An example of XGBoost's success can be given by KDDCup 2015 \citep{KDDCup2015}, where all teams in the top 10 used it. To enable the application of XGBoost, this study introduces a compact feature-based representation of the original image-based remote sensing data. Results are evaluated based on an extensive experimental evaluation on the annual soybean yields within the United States, captured and provided by the U.S. department of agriculture (USDA) \citep{USDA}. The XGBoost-based approach for yield prediction is compared with two state-of-the-art approaches to yield prediction based on Deep Learning. The comparative evaluation shows that the XGBoost-based approach is at least on par with both Deep Learning-based approaches. Lastly, trust in the model is increased by using a feature importance analysis based on Shapley Values to showcase that the important features within the model are in line with traditionally used inputs for crop monitoring, most importantly the reflection of light in the near-infrared spectrum.
\par
The results help to enable the real-world adoption of Machine Learning assisted yield prediction. On the one hand, high accuracies are important for meaningful predictions. On the other hand, utilizing a well understood method such as XGBoost makes for a trustworthy prediction. Future work will need to monitor whether larger training datasets available in the future will increase the accuracy of the Deep Learning approaches. Consequently, we aim to perform experiments using our approach on different crops and in different regions to demonstrate a wide area of application.

\subsection{Contributions}
The contributions of this study are the following:
\begin{itemize}
\item System: Describing a complete processing pipeline to apply XGBoost for yield prediction based on a time series of remote sensing data. (Section \ref{sec:mat and met})
\item Effectivity: Extensive comparative experiments on soybean yield data in the United States are conducted, hinting at the capabilities of \mbox{XGBoost} on yield prediction by outperforming two state-of-the-art approaches for yield prediction based on Deep Learning by an average of at least 25\% in terms of $\operatorname{RMSE}$. (Section \ref{experimental results})
\item Explainability: Shapley Value feature importances are utilized to raise trust in the model and prove that the model's important features are in line with expert knowledge in the field of yield prediction. (Section \ref{sec:explainability})
\end{itemize}

\subsection{Related Work}
\label{sec:related}

A variety of Machine Learning methods can be encountered that support automated approaches to crop yield prediction. Several comprehensive surveys on automated approaches to crop yield prediction are given by \cite{van2020crop} and \cite{TurkishSurvey2021}. 

For this study, the discussion is focused by dividing the research field into three types of approaches to yield prediction. First, solutions based on statistical models. Second, classic Machine Learning approaches and (3) solutions based on Deep Learning. In addition, the advances that other studies have made in explaining their predictions are highlighted.  

\subsubsection{Statistical Models for Yield Prediction}
Two representative approaches to the prediction of yield based on statistical models and vegetation indices are reported by \cite{meng2019assessment} and \cite{zhao2020predicting}.  \cite{meng2019assessment} propose a new daily vegetation index based on MODIS data and use it to predict cotton yield in California. An exponential function with two learned parameters is used to predict the yields. They achieve good results for yield prediction on a field scale by fusing imagery with low and high spatial resolutions. \cite{zhao2020predicting} predict wheat yields at a field scale in Wales with traditional crop modeling based on high-resolution images from the Sentinel-2 satellite. Their yield prediction model was developed using statistical analysis of variance (ANOVA) and a multivariate analysis that incorporates various derived metrics and indices.
\par
While these approaches are data- and computationally efficient, the hand-crafted modeling of multiple complex interactions with statistical models is very rigid and vulnerable to not scaling well with changes in the environment. Therefore, the XGBoost-based approach within this study aims to benefit from the intrinsic advantages of Machine Learning approaches, i.e., being adaptable to changes in the environment by retraining using new training data. 

\subsubsection{Classic Machine Learning for Yield Prediction}

The works of \cite{rodriguez2017machine} and \cite{bobeda2018using} focus on the prediction of yields in citrus orchards. Both approaches use a selection of phenological information that is captured directly from the plants multiple times a year. For both, the M5-Prime \citep{wang1996induction} implementation of Random Forests shows superior results on different kinds of citrus fruits. \cite{sirsat2019machine} describe an approach to yield prediction for grapevines using phenological information, soil properties, and climatic conditions. The Random Forest ensemble technique proposed by \cite{breiman2001random} shows the best results for predicting grapevine yield.

More recent publications show the benefits of incorporating remote sensing data for yield prediction. \cite{martinez2020crop} include remote sensing data in terms of vegetation indices together with phenological information and use a Gaussian Process to predict the yield of corn, wheat and soybeans in the United States. \cite{meng2021predicting} employ vegetation indices for the prediction of maize yield in California on a field scale. Complementing data sources include fertilizer information, climate data, and soil properties. They conclude that non-linear models such as Random Forests are best performing for this kind of prediction task.

The first study to use XGBoost for yield prediction is that of \cite{charoen2018sugarcane}. Based on plot characteristics, phenological information and rain volume, they predict the yield of sugar plots in Thailand. This approach forgoes an exact yield regression and instead solves the binary classification problem, whether the yield is above or below the median of the training data. \cite{cao2020identifying} deploy LightGBM as a form of Gradient Boosting for Random Forests to predict winter wheat yields across China. They again use vegetation indices, derived from remote sensing data, together with climatic and socioeconomic factors as input data.

\subsubsection{Deep Learning for Yield Prediction}
\label{sec:deeplearningforyieldprediction}
The extensive use of Deep Learning for the prediction of yields gained momentum with the publication of \cite{you2017deep}. They investigated the deployment of basic Convolutional Neural Networks (CNNs) and Long short-term memory Networks (LSTMs) for yield prediction on a USDA soybean dataset similar to the one used in this study. Furthermore, they developed the idea of using binned histogramizations of the remote sensing data to overcome the limitations that are implied by nonregular sized images for Deep Learning. The same concept is picked up by \cite{sun2019county}, who evaluate a CNN-LSTM hybrid model for the same yield prediction task and report improved accuracy. Other notable publications, including CNNs, are presented by \cite{wang2020winter} and \cite{wolanin2020estimating}.
\par
Forgoing complex structures such as CNNs and RNNs, the work of \cite{khaki2019crop} shows that an accurate prediction of maize yield based on environmental and genotype information is possible by using solely simple Feed Forward Networks. They predict future weather conditions to obtain a more accurate prediction of yield in the early stages of the year.
\par
\cite{alibabaei2021crop} demonstrate the benefits of incorporating the exact irrigation schedule of the examined crops. Given very detailed input information, they compare predictions made by several algorithms including LSTMs, CNNs and Feed Forward Networks. The experiments show a superior performance of a bidirectional LSTM structure in predicting the yields of potatoes and tomatoes.
\par
Data provided by the USDA has become very popular in recent years because it is one of the largest sources of yield data. Like most regression tasks, yield prediction becomes more difficult when there are less training data available. \cite{wang2018deep} address this problem by applying the concept of transfer learning. They first train an LSTM network for soybean yield prediction in Argentina and then transfer the knowledge to regions in Brazil that have fewer available data. Another idea to solve this problem is introduced by \cite{khaki2021simultaneous} who combine the predictions of different crops to overcome data scarcity. They present a multi-target regression CNN structure, together with a newly developed combined loss function. Experiments are again conducted on the USDA data by simultaneously predicting soybean and corn yields based on remote sensing data, showing promising results.
\par
There are two state-of-the-art approaches to soybean yield prediction based on deep learning that are explicitly demonstrated and evaluated on USDA and remote sensing data, namely the CNN-based approach of \cite{you2017deep} and the CNN-LSTM hybrid model of \citet{sun2019county}. 
Therefore, the approach of an XGBoost-based yield prediction system will be compared with these two approaches (Section \ref{experimental results}). 

\subsubsection{Explainability in Yield Prediction}
While the accuracy of yield prediction increases, it is not the only aspect of importance within the research area. A detailed understanding of the model and the impact of the different inputs is also important. In general, this increases trustworthiness of the model to the user-base and can help to understand the complex relations determining the yearly yields.

Since data acquisition in the context of yield prediction is usually expensive, most studies focus on finding important subsets of features within the entirety of their data. A commonly utilized method is to create distinct datasets that each include a different subset of features. The entire training and evaluation process is executed with each dataset individually, and the resulting accuracy is associated with the importance of the selected features. This is done by \cite{bobeda2018using} and \cite{cao2020identifying} to predict the yields of citrus fruits and winter wheat, respectively. The studies reveal that vegetation indices and environmental effects, such as climatic conditions, are important to explain yields. For citrus fruits, especially, basic information about the trees, such as their age, was shown to be important.

Another common approach to add explainability to yield prediction is to analyze feature importances inherent in some Machine Learning algorithms. As stated above, the work of \cite{sirsat2019machine} and \cite{meng2021predicting} includes Random Forests for the prediction of the yield. In both works, they analyze feature importances based on Random Forest's inherent structure in order to draw conclusions about their model. \cite{sirsat2019machine} show a high importance of climatic conditions as opposed to other data sources, such as fertilizer and soil properties, for predicting grapevine yields. \cite{meng2021predicting} show a high importance of remote sensing data in form of vegetation indices for the prediction of maize yields. As another instance of built-in feature importances, \cite{martinez2020crop} interpret the optimization parameters learned within the Gaussian Process as indicators for each feature's relevance within the model. The results show a high importance of features associated with vegetation indices when predicting different field crops within the United States. 

For Deep Learning approaches, no built-in importance measure can be used for explainability. \cite{you2017deep} overcome the ``black-box'' nature of Deep Learning by randomly permuting the values of specific features within the entire dataset before training and testing their model. Since they work solely on remote sensing data, they perform this procedure either for the values of a specific band or for the values of specific time steps. For predicting soybean yields within the United States, they conclude that the near-infrared band is important, which corresponds to the consensus in traditional crop monitoring. \cite{khaki2019crop} solve the problem of explainability for Deep Learning approaches by applying backpropagation to the gradients of neurons of the last layer of their network. They select the neurons that are on average very active across the entirety of the validation data and use backpropagation to find the input features that are most influential to the neurons being active. For the prediction of maize yield, they find that environmental effects are more important than genotype information. 

\section{Material}
\label{sec:mat and met}
The data chosen to evaluate different yield prediction approaches are the soybean production reported within the United States over a time frame of 2003 to 2021. This allows good comparability with previous work, since soybeans in the United States are widely investigated \citep{you2017deep,sun2019county}.

\begin{figure}
	\centering 
	\includegraphics[width=\textwidth]{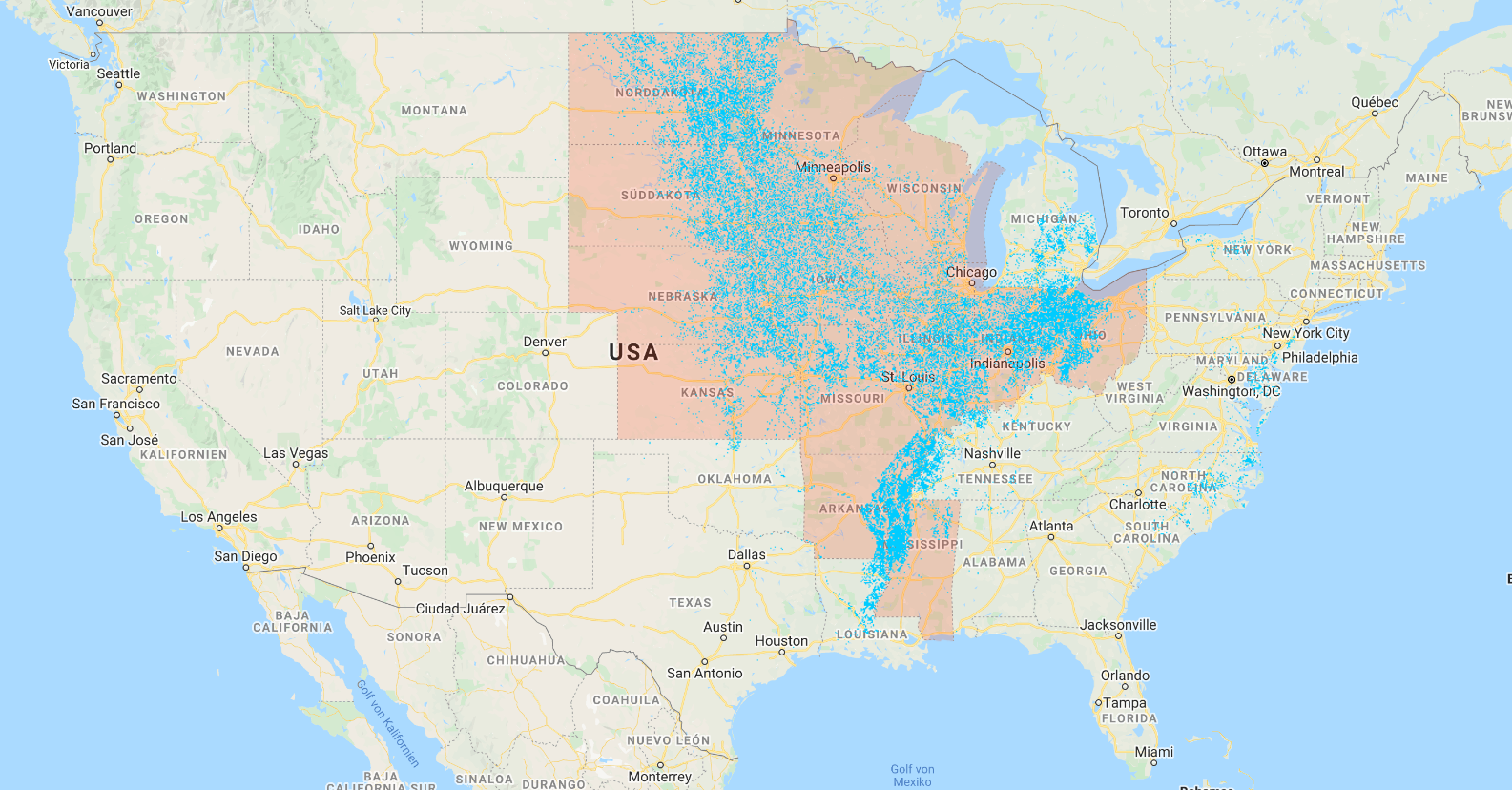}
	\caption{The 13 adjacent states of the United States included in this study are highlighted in red. The blue areas indicate soybean farmland. The image is extracted via Google Earth Engine \citep{gorelick2017google}.}
	\label{fig:13states}
\end{figure}

The study area includes 13 adjacent states within the United States. The states are Arkansas, Illinois, Indiana, Iowa, Kansas, Minnesota, Mississippi, Missouri, Nebraska, North Dakota, Ohio, South Dakota and Wisconsin. The 13 states are highlighted in Figure \ref{fig:13states}, showing that most of the soybean farmland in the United States is located within the selected states. 

The study spans across 19 years, from 2003 to 2021. For the end-of-year soybean yield prediction, predictions are based on data captured over the timespan between the 49th and 321st days of the year. Since the soybean cycle varies between years and state selection, generous ranges around the usual dates for planting and harvesting are used. The range includes approximately two months before the first crops are planted and ends approximately a week after the last soybeans are harvested \citep{USDA_dates}. 

The input features for the prediction of yield are extracted from five different data sources that have also been used to predict soybean yield by \citet{sun2019county}. 

\subsection{USDA Yield Data}
\label{sec:USDAYield}
County-level soybean yield data from 2003 to 2021 are collected by the USDA \citep{USDA}. The soybean yields are reported in soybean bushels per acre (bu/ac) with 1 bu/ac being approximately equivalent to 67.26 kg/ha.
The yield data are used as ground truth labels for model training and validation.

\begin{figure}
	\centering 
	\includegraphics[width=\textwidth]{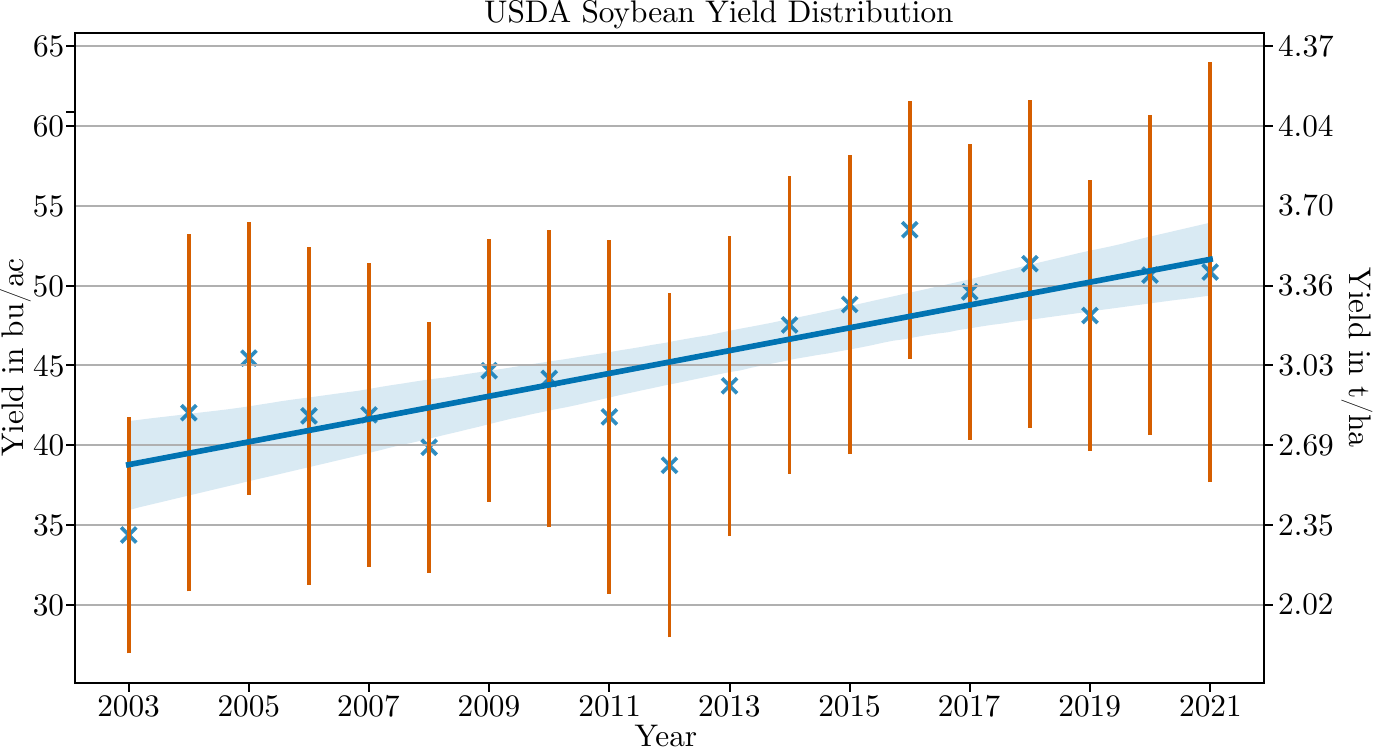}
	\caption{The historic yields for the study time frame. The crosses indicate the mean yields in bu/ac (t/ha) with the yellow bars describing the standard deviations. The blue line depicts a trend line, together with a confidence interval.}
	\label{fig:mean}
\end{figure}

Figure \ref{fig:mean} depicts the historic yields during the study period, and the graph shows an overall upward trend, caused mainly by technological innovations leading to more efficient farming \citep{chambers2020sources}. 

\subsection{MODIS Surface Reflectance and Temperature}
\label{sec:modis}
Within this study, remote sensing data collected by Moderate-Resolution Imaging Spec\-troradiometers (MODIS) installed upon the NASA Terra and Aqua satellites are used. The MODIS data are available from the Google Earth Engine Catalog \citep{gorelick2017google} and consist of the MODIS products MOD09A1 that provide surface spectral reflectance data and MYD11A2 which provide land surface temperature data. 

In this study, both products are used at a resolution of 500 m per pixel, with each image being an 8-day composite of the values of each band. The surface reflectance consists of 7 bands covering visible and non-visible spectrums of light. The visible spectrum includes blue, green, and red light, which can be used, e.g., for RGB visualizations of the regions. The non-visible spectrums are 4 different wavelengths of infrared light and near-infrared light. The importance of near-infrared light for yield prediction is well known and is utilized in standard measures such as the Normalized Difference Vegetation Index (NDVI) \citep{gao1996ndwi}. A temperature image shows two bands, one for the temperature during the day and one for the temperature during the night. Using the timespan between the 49th and 321st days of the year therefore produces 34 images to correspond to one year's yields. 

\subsection{Daymet Weather Data: Precipitation and Vapor Pressure}
\label{sec:daymet}
The Daymet V4 dataset offers daily surface weather and climatological summaries extracted from the Google Earth Engine Catalog \citep{thornton2016daymet}. It provides gridded estimates of daily weather parameters based on selected meteorological station data and various supporting data sources within the United States. Two important weather parameters in Daymet — daily total precipitation and daily averaged partial pressure of water vapor — produced on a 1 km $\times$ 1 km gridded surface over North America were selected as climatic factors as proposed by \cite{sun2019county}. The scaling is changed from 1000 m to 500 m by using the native Earth Engine image pyramid to match the weather data with the other yield and environmental data. Lastly, the daily data is aggregated to match the same 8-day composites of the MODIS data by applying the mean. 

\subsection{Tiger County Borders}
\label{sec:Tiger}
The United States Census Bureau TIGER dataset contains the 2018 boundaries for the primary legal divisions of the US states \citep{Tiger}. The information is used to crop the remote sensing data to the county borders.

\subsection{USDA NASS Cropland Data Layer}
\label{sec:CDL}
The Cropland Data Layer (CDL) is a crop-specific land cover data layer created annually for the continental United States and also downloaded from the Google Earth Engine catalogue. The selection of 13 adjacent states within the United States has been partly covered since 2003 and fully covered since 2006. The states Kansas, Mississippi, Missouri, Ohio, and South Dakota are the ones additionally covered since 2006, when the CDL started to be available for the entire United States. Originally, the CDL has a resolution of 30 m. This resolution of 30 m is upscaled to a resolution of 500 m to match the resolutions of the other data sources explained above. The CDL information is used to focus the training of the prediction system to the parts of the remote sensing images that cover soybean farmland.

\section{Methods}
\label{sec:data preparation}
\begin{figure}[ht]
	\centering 
	\includegraphics[width=\textwidth]{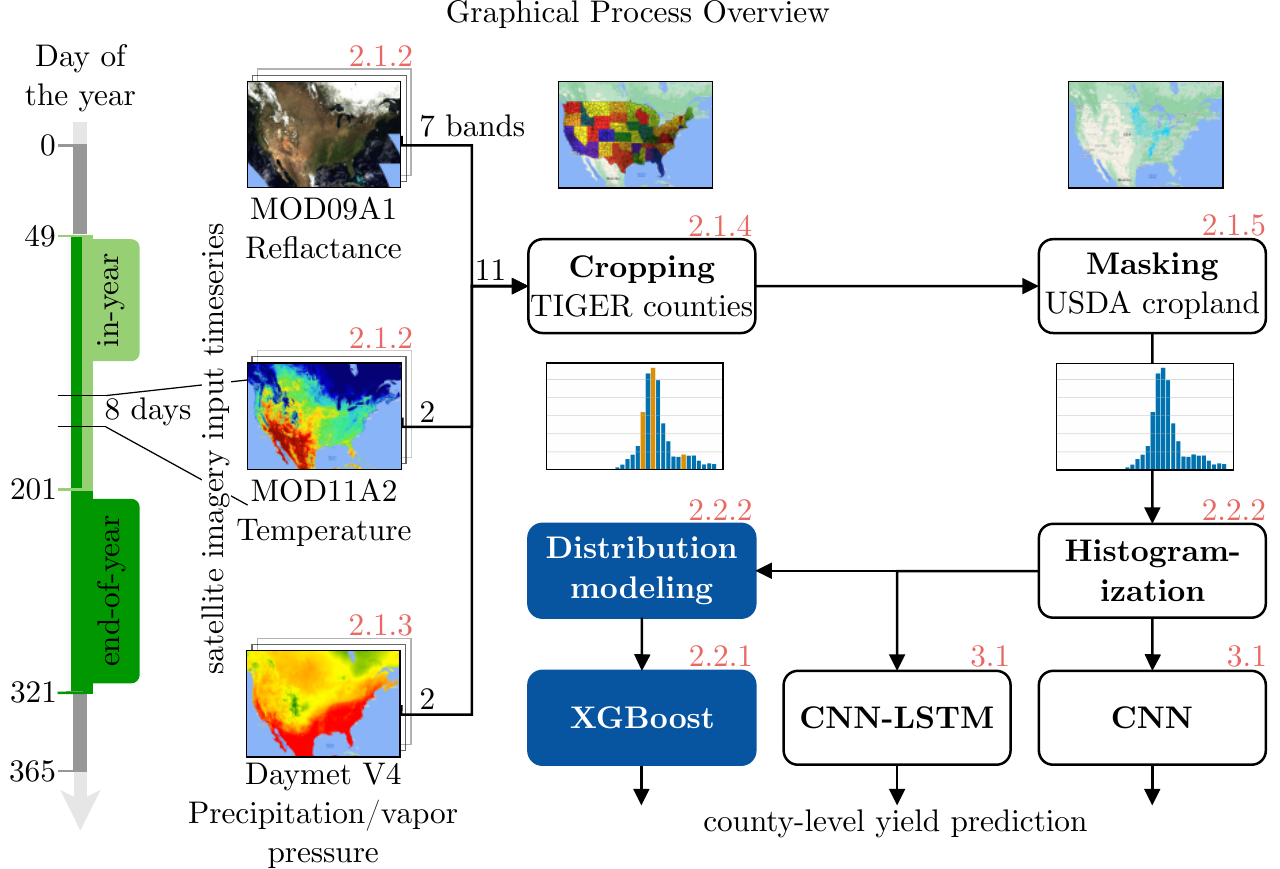}
	\caption{Overview of the remote sensing data processing pipeline for yield prediction. The aggregated remote sensing data are cropped and masked, before they are processed. The resulting data are then either represented by histograms for the Deep Learning approaches or a feature-based approximation of the underlying distribution for XGBoost.}
	\label{fig:overview}
\end{figure}
A graphical overview of the complete prediction pipeline is given in Figure \ref{fig:overview}. The process starts with determining the time window that is used for the predictions. Either the in-year prediction between day 49 and day 201 of the year or the end-of-year prediction between day 49 and day 321 of the year is performed. As explained above in Section \ref{sec:mat and met} three different satellite imagery products are utilized to build the input data for the pipeline. Namely, the MOD09A1 reflectance data, the MOD11A2 temperature data (Section \ref{sec:modis}) and the DaymetV4 precipitation and vapor pressure data (Section \ref{sec:daymet}). The relevant sections for the prediction of yield at the county level are obtained using the Tiger county borders (Section \ref{sec:Tiger}) to crop the data and by applying the USDA cropland data (Section \ref{sec:CDL}) as a mask to keep only pixels covering soybean farmland.

The experimental setup focuses on a direct comparison between XGBoost-based yield prediction and current state-of-the-art Deep Learning frameworks. After the cropping and masking explained above, the resulting images have irregular shapes and varying sizes between different counties. Furthermore, the preferred presentation of data as input for Deep Learning frameworks and classical Machine Learning approaches differs. Therefore, the data are either processed towards a histogram-based approach for Deep Learning or a feature-based modeling of the underlying pixel distribution for the \mbox{XGBoost} approach. The histogramization will output the relative frequency of the pixel values in 32 evenly spaced intervals, and the distribution modeling will highlight three key characteristics of the underlying distribution (Section \ref{sec:hists_and_params}). Lastly, the different approaches are applied on their respective preprocessed data sources to compute both the in-year and the end-of-year predictions.

\subsection{Yield Prediction with Gradient Boosting}
\label{sec:XGBoost}

Among the Machine Learning methods used in practice, gradient tree boosting (GTB) or gradient boosted regression tree (GBRT) and especially, the XGBoost approach of \citet{chen2016xgboost} has been shown to give state-of-the-art results in many standard classification and regression benchmarks and competitions. For example, in KDDCup 2015 \citep{KDDCup2015}, all teams in the top 10 used XGBoost.

A single regression tree is a set of cascading questions. Applied to a data point (i.e., a set of features and their values), the feature values are used to answer the cascading questions, yielding a result value. Regression trees are trained on training data to learn the appropriate cascade of questions.

A tree ensemble is a family of $K$ regression trees where the final prediction is the sum of the predictions for each tree. XGBoost learns the appropriate tree ensemble from training data by iteratively adding new trees into the given ensemble that maximize the prediction performance of the trees already in the ensemble \textit{plus} the new tree that has to be added. 
In other words: for a given tree ensemble $E_k$ of $k$ trees and its residuals, a new tree $e_{k+1}$ that fits best to these residuals is combined with the given ensemble $E_k$, which produces the boosted version $E_{k+1}$ of $E_k$. The performance of $E_{k+1}$ will be better than that of $E_k$. This can be done for $K$ ($1 \le k \le K$) iterations, until the residuals have been minimized as much as possible. Therefore, the design of the new tree uses the gradients of the performance evaluation functions to select the best cascade of questions.

\subsection{XGBoost Regression Design}
To extend the general explanation of the XGBoost algorithm above, a more in-depth explanation of the XGBoost regression design is given in this section. In general, the following objective function needs to be optimized:
\begin{equation}
\label{eq:generalobjective}
O(M)=L(M)+\Omega(M),
\end{equation}
where $M$ is the model learned from the training data. $L$ is a differentiable convex training loss function, such as the Mean Squared Error (MSE) and $\Omega$ is the regularization term which prevents overfitting, i.e., the model learning the structure of the training data too precisely, leaving it without the ability to generalize to unseen data. 

As explained above, the model is an ensemble of decision trees, so the output $\widehat{y}_i$ for each instance $i$ is given by a collection $E_K$ of $K$ trees. 

\begin{equation}
\widehat{y}_{i}=\sum_{k=1}^{K} e_{k}\left(x_{i}\right), e_{k} \in E_K .
\end{equation}

Since the model in the general objective \eqref{eq:generalobjective} includes the decision trees as functional parameters, the loss cannot be optimized directly, but it has to be done iteratively, where in the $t$-th step the following is minimized:
\begin{equation}
\label{eq:objective}
O^{(t)}=\sum_{i=1}^{n} L\left(y_{i}, \widehat{y}_{i}^{(t)}\right)+\sum_{k=1}^{t} \Omega\left(e_{k}\right),
\end{equation}

where $n$ is the number of training data points, and for a given iteration, $\widehat{y}_{i}^{(t)}$ can be calculated as:
\begin{equation}
\widehat{y}_{i}^{(t)}=\sum_{k=1}^{t} e_{k}\left(x_{i}\right)=\widehat{y}_{i}^{(t-1)}+e_{t}\left(x_{i}\right),
\end{equation}
by adding the prediction of the latest calculated tree to the previous prediction.

\citet{chen2016xgboost} define the regularization term $\Omega\left(e_{k}\right)$ as:
\begin{equation}
\Omega\left(e_{k}\right)=\gamma T+\frac{1}{2} \lambda \sum_{j=1}^{T} w_{j}^{2},
\end{equation}
Where $T$ is the number of leaves in the tree penalized by $\gamma$ and $w_j$ are the leaf weights within the tree penalized by the parameter $\lambda$. The leaf weights represent the answer of the tree that is given for the regression task, when the cascade of questions reaches the particular leaf. Lower numbers of leaves with smaller leaf weights force the tree to learn a very generalized representation of the training data. 

To transform the problem of optimizing the objective function into a simpler problem of minimizing a quadratic function, \cite{chen2016xgboost} perform multiple steps. They apply the second order Taylor approximation and perform further steps to obtain the following approximate representation of the objective function \eqref{eq:objective}:
\begin{equation}
\label{eq:final_objective}
O^{(t)} \approx \sum_{i=1}^{n} \frac{1}{2} h_{i}\left(e_{t}\left({x}_{i}\right)-g_{i} / h_{i}\right)^{2}+\Omega\left(e_{t}\right)+c, 
\end{equation}
where $g_i$ and $h_i$ are the first and second order derivatives of the MSE loss function and $c$ is a constant term. The new representation of the objective function can be used to find the optimal weights $w_i$ that minimize the loss function at each step.

The last remaining problem is to find the optimal cascade of questions to ask for each tree. More precisely, this refers to finding the features and values to define the splits at each node of the tree. \cite{chen2016xgboost} propose two algorithmic solutions to the problem. First, an exact greedy algorithm to solve this problem by enumerating all possible splits and evaluating each option with the help of equation \eqref{eq:final_objective} and secondly an approximate algorithm. The approximate algorithm is chosen for this study since it is best suited to handle the amount of training data within the experiments. The basic idea of the approximate algorithm is to find promising possible splittings based on the given feature distributions, resulting in improved time efficiency by not having to evaluate all possible splits. 

Lastly, the clever system design of \cite{chen2016xgboost} further differentiates between classical gradient boosting and XGBoost. They introduce the concept of \textit{blocks} in their work. Because data need to be put in order multiple times during the approximate split finding algorithm, data are stored in blocks, where each column is already sorted by the corresponding feature value. The blocks can be divided and used for parallel computing of the approximate split finding algorithm.

\subsection{Hyperparameter Tuning}
\label{sec:hyper}
When applying XGBoost to a specific dataset, a multitude of hyperparameters is able to be adjusted to achieve the highest possible accuracy. A Tree-structured Parzen Estimation (TPE) \citep{bergstra2013making} is deployed as a sequential model-based optimization approach. This means that models are sequentially constructed to approximate the performance of the model for a set of hyperparameters based on historical measurements. TPE tries to estimate the underlying relations between a quality measure and the hyperparameters by exposing the underlying expression graph of how a performance metric is influenced by the hyperparameters. For this study, the Python implementation of TPE within the Optuna framework is used for the experiments \citep{akiba2019optuna} and ran for 50 iterations without further pruning involved to optimize the accuracy on the validation set. A model is tuned and trained for each testing year from 2017 to 2021, using 10\% of the training data available for validation. After being used for validation, the data is included again in the training set for the final model. A unique model is trained for each year, to be as close as possible to the real-world use case, where it should be beneficial to include the most recent years in training.

\subsection{From Raw Images to Histograms and Feature-Based Representations}
\label{sec:hists_and_params}
The downloaded satellite images form input sequences $\mathbf{d} \in \mathbb{R}^{34 \times H \times W \times 11}$, where $H$ and $W$ are the variable height and width of the images, 34 is the number of 8-day episodes and 11 is the number of bands. $H$ and $W$ are variable throughout the data, as the size of the counties varies. 

Depending on whether an end-of-year prediction with information available covering the whole growing period or an in-year prediction with information available to a certain point is performed, the corresponding time frames of the remote sensing data are selected. The satellite images are cropped to the county borders and masked with the CDL. Pixels showing non-zero values are distributed unevenly. A common workaround is to process each band of the images into histograms with evenly distributed bins \citep{you2017deep}. This approach reduces the data size dramatically and still conserves sufficient information for Deep Learning approaches to learn expressive representations of the data. 

However, since the bin values of the histograms are standardized over all images on the one hand and pixel values vary significantly between different images on the other hand, the resulting histograms show bins filled with zero valued entries. 

\begin{figure}
	\centering 
	\includegraphics[width=\textwidth]{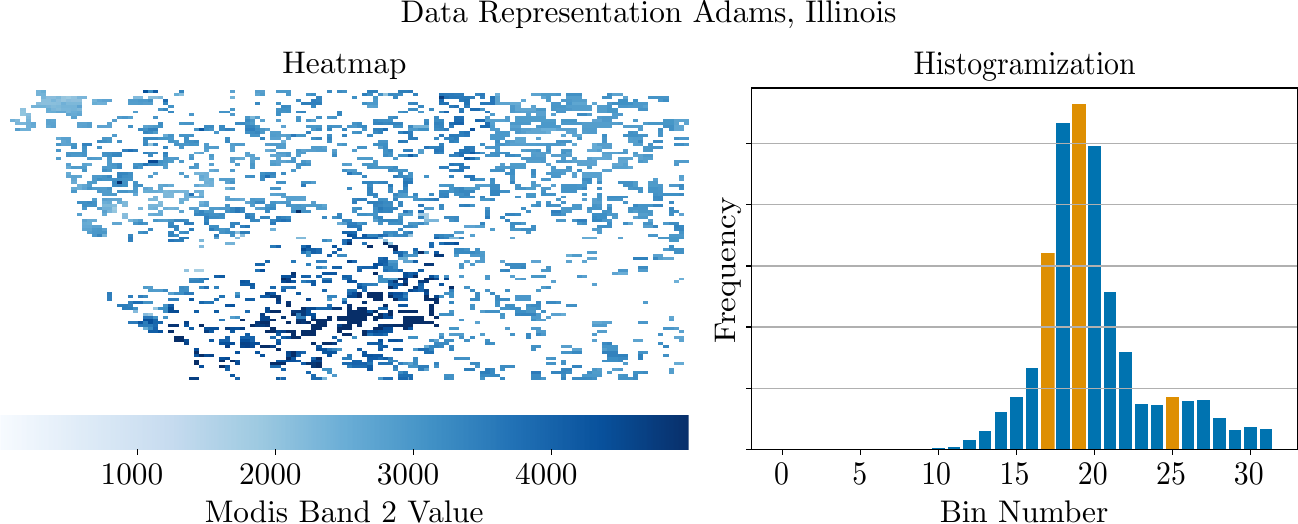}
	\caption{Left: Heatmap for band 2 of the masked MODIS surface reflectance data, captured between day 105 and day 113 of 2020 in Adams, Illinois. Right: The histogram showing the relative frequencies of the same data. The bins depicting the 20\% quantile, the median and the 80\% quantile are highlighted in orange.}
	\label{fig:combined2}
\end{figure}

 The value distributions of the satellite bands mostly follow skewed normal distributions, allowing for an approximation of the value distribution of each band of an image by a normal distribution parameterized by three values, namely the median and the values marking the 20\% and 80\% quantiles. The choice of features is well suited to describe the skewed normal distributions. While the median is an outlier resistant approximation of the expected value of the distribution, a small and a high quantile value are useful to monitor the slopes to the left and to the right of the median. Experiments showed that the specific values 20\% and 80\% are a good choice for the task. This is the feature-based representation used as input for XGBoost.
Intuitively, it is possible to model remote sensing data as skewed normal distributions, since soybean acres within proximity to each other will show similar surface reflections and are exposed to similar weather conditions. An example is shown in Figure \ref{fig:combined2}. The sparse heatmap depicts the initial data. The orange bars within the histogram representation show how the three selected values describe the underlying skewed normal distribution.
\par
It is important to note that the proposed reduction in dimensionality is based on the assumption of permutation invariance. This means that the prediction of future yields relies more on the value of the non-cropped pixels than on their location. As \cite{you2017deep} point out, this assumption ignores the likely dependencies of the target output on position-bound features such as soil properties or elevation. \cite{you2017deep} therefore combined their Deep Learning approaches with a Deep Gaussian Process to integrate the missing spatio-temporal information, resulting in only a small improvement of 3.9\% in terms of $\operatorname{RMSE}$. This must be related to the significant reduction in dimensionality by removing the spatial information from the data (see Section \ref{sec:data_and_time_efficiency}). 
\par
Finally, all the values of every band from one sequence are concatenated. Multiplying the 3 values of the feature-based distribution representation of all 11 bands over 34 time steps results in an initial input vector $x \in \mathbb{R}^{1122}$. Since classic Machine Learning models are able to take advantage of handcrafted features, another seven features are added. Positional awareness is added to the prediction by including the latitude and longitude of the respective centers of the counties. Furthermore, the awareness of time is raised by including the year, the sample is taken from, and the number of years passed since 2003. The county-wise average yield computed on the previous years is added to incorporate spatial dependencies. Lastly, the y-intercept and the slope of a linear regression fitted to the historic yields in the dataset are included. These features help to explain a growing trend in average soybean yields over the years.

\subsection{State-of-the-Art Approaches to Soybean Yield Prediction}
The XGBoost-based approach is compared with two Deep Learning approaches to soybean yield prediction. 
The CNN network presented by \cite{you2017deep} and the CNN-LSTM hybrid network structure presented by \cite{sun2019county}. Both approaches convert the original satellite images into image histograms of each band with evenly distributed bins, as described in Section \ref{sec:hists_and_params}. 

 For the histogramization of the bands provided by the MODIS products, the data processing follows the limits given by \cite{you2017deep}. The original work did not use the Daymet data, but in this study they are included in both approaches for a fairer comparison. Furthermore, the theoretical minimal and maximal values are used as the boundaries for the histograms of the bands provided by the Daymet product. The hyperparameters described in the respective work for both networks are used, including an early stopping criterion when the validation score shows no improvement for 10 epochs. The results are averaged over five runs to account for randomness. The implementation of the CNN network is taken from \citep{github} and the CNN-LSTM hybrid approach from \cite{sun2019county} is implemented from scratch. As for XGBoost, a new model is trained for each year with access to information across all previous years. A random 10\% split of the training data are used for validation and all data are normalized by substracting the mean of the training data.
 
 The CNN structure of \cite{you2017deep} is very straightforward, consisting of seven convolutional layers of varying sizes with either stride-1 or stride-2, followed by a fully connected layer with 2048 neurons for the final prediction. The stride-1 layers serve the purpose of a classical convolution. The stride-2 layers are important to reduce dimensionality throughout the network, since no pooling layers are used.
 
 The CNN-LSTM network of \cite{sun2019county} applies a CNN structure to the input data to learn a meaningful representation that is used as input for a standard LSTM network. The CNN structure consists of two blocks of convolution, batch normalization, and max pooling that are concatenated and used for each of the 34 time steps separately. The output is flattened and serves as an input for the LSTM network with a hidden layer size of 256. Lastly, the output of every iteration is passed through a fully connected layer with 64 neurons and concatenated with each other. The final result is calculated by utilizing a dropout layer with a probability of 0.5 before using a last fully connected layer for the singular output.

\subsection{Performance Metrics}
For the evaluation, two main metrics to quantify prediction performance are used. To measure the overall error, the Root Mean Square Error ($\operatorname{RMSE}$) is calculated as follows:
\begin{equation}
       \operatorname{RMSE}(\hat{\mathbf{y}}, \mathbf{y})=\sqrt{\operatorname{MSE}(\hat{\mathbf{y}}, \mathbf{y})}=\sqrt{\frac{1}{n} \sum_{i=1}^{n}\left(\hat{y}_{i}-y_{i}\right)^{2}}.
\end{equation}
Here, $\mathbf{y}$ represents the ground truth values and $\hat{\mathbf{y}}$, the prediction values. To be able to measure the variability in the target variable that is explained by the models, the coefficient of determination ($\operatorname{R^2}$) is utilized:
\begin{equation}
\operatorname{R^2}(\hat{\mathbf{y}}, \mathbf{y}) = 1-\frac{\sum_{i=1}^{n}\left(y_{i}-\hat{y}_{i}\right)^{2}}{\sum_{i=1}^{n}\left(y_{i}-\bar{y}\right)^{2}},
\end{equation}
where $\bar{y}$ is the mean of the ground truth values $\mathbf{y}$.

Furthermore, every experiment consists of five runs to account for randomness in initialization of the competing Deep Learning frameworks and sub-sampling within the XGBoost algorithm.

\section{Experiments and Discussion}
\label{experimental results}
The performance of the XGBoost-based approach for yield prediction and the Deep Learning approaches are evaluated in two different scenarios. An end-of-year prediction and an in-year prediction on soybean yields provided by the USDA.

\subsection{End-of-the-Year Prediction}

Experiments are conducted on the entire dataset to evaluate all approaches for an end-of-year yield prediction. This means the complete span between the 49th and 321st day of the year is used. By this date, all soybeans should be harvested \citep{USDA_dates}, and therefore the satellite data includes information on the farmland right before harvest. When testing the approaches for a specific year of the dataset, it is assumed that only data from before the specific year is available for training and validation, while the year itself is excluded explicitly for testing.

The results can be observed in Table \ref{table:end1}. When comparing the XGBoost-based approach to the Deep Learning approaches, the $\operatorname{RMSE}$ decreases by about 25\% when averaging the metric over the five years in the testing data. In the same time frame, the average $\operatorname{R^2}$ score increases by about 0.13. The good performance of the XGBoost-based approach in the experiments can be explained by the nature of the data. Although remote sensing data are often referred to as satellite images, there is a higher resemblance to tabular datasets, where the information is represented by numerical features instead of pixels. The necessary preprocessing towards histograms or distribution approximation values strengthens this supposition. Classic Machine Learning approaches on feature engineered datasets thrive under these conditions.

Regarding the results of the hyperparameter tuning as explained in Section \ref{sec:hyper}, a tendency of the models to develop deep trees is prevalent. The parameter that provides the maximum allowed depth of the individual trees averages 21.8 over all five years in the experiments. The XGBoost library defaults towards a value of 6 for this specific parameter, since the standard use cases contain smaller quantities of overall features. This indicates that the data at hand are of high complexity and need a lot of features to be included for an informed regression. Since this could lead to overfitting, other hyperparameters are tuned to encourage conservative model behavior. The learning rate is rather low and averages at 0.06 compared to the default value of 0.3. Furthermore, the hyperparameters indicating stochastic sampling of rows and columns in each iteration average at 0.6 and 0.84 respectively, to help avoid overfitting. 

\begin{table}
\begin{tabularx}{\textwidth}{ |c| *{6}{Y|} }
\cline{2-7}
   \multicolumn{1}{c|}{} 
 & \multicolumn{2}{c|}{XGBoost}  
 & \multicolumn{2}{c|}{CNN}
 & \multicolumn{2}{c|}{CNN-LSTM}\\
\hline
 Year & $\operatorname{RMSE}$ & $\operatorname{R^2}$ & $\operatorname{RMSE}$ & $\operatorname{R^2}$ & $\operatorname{RMSE}$ & $\operatorname{R^2}$\\
\hline
2017 & {\leavevmode\color[HTML]{009901}3.77}  & {\leavevmode\color[HTML]{009901}0.82} & {5.01}       & 0.70      & {5.07}         & 0.70         \\ 
2018 & {\leavevmode\color[HTML]{009901}4.51} & {\leavevmode\color[HTML]{009901}0.76} & 6.15 & 0.63 & 6.21 & 0.63 \\ 
2019 & {\leavevmode\color[HTML]{009901}4.21} & {\leavevmode\color[HTML]{009901}0.76} & 5.52 & 0.57 & 5.76 & 0.54 \\ 
2020 & {\leavevmode\color[HTML]{009901}4.22} & {\leavevmode\color[HTML]{009901}0.80} & 6.66 & 0.55 & 7.63 & 0.42 \\ 
2021 & {\leavevmode\color[HTML]{009901}4.55} & 0.82 & 5.12 & {\leavevmode\color[HTML]{009901}0.85} & 6.03 & 0.79 \\ 
AVG  & {\leavevmode\color[HTML]{009901}4.25} & {\leavevmode\color[HTML]{009901}0.79} & 5.69 & 0.66 & 6.14 & 0.61 \\
\hline
\end{tabularx}
\caption{\label{table:end1} $\operatorname{RMSE}$ and $\operatorname{R^2}$ values of \textbf{\textit{end-of-year soybean yield predictions}}. The best result for every year is highlighted in green. The CNN architecture is taken from \citet{you2017deep}, and the CNN-LSTM architecture is taken from \citet{sun2019county}.}
\end{table}

\subsection{In-Year Prediction}
To showcase the dynamics of the in-year soybean yield estimation, the time between the 49th and the 201st day of the year is used to produce the training data. This ensures that the harvesting season has not yet started. Therefore, the data contain 19 of the 8-day intervals that present the remote sensing data. The reduced amount of information increases the difficulty of the prediction task. On the same note, information gained from an in-year prediction is even more valuable to every party involved, since classic estimation approaches often include counting specific features of the crop that might not have developed yet. Training and testing data are handled as explained for the end-of-year prediction.

The results are presented in Table \ref{table:mis1}. Averaging the error metrics over the five testing years, the XGBoost-based approach outperforms the best state-of-the-art approach by 36\% in terms of $\operatorname{RMSE}$ and 0.42 in terms of $\operatorname{R^2}$. The gap between the Deep Learning approaches and the XGBoost approach appears to be wider than for the in-year prediction. This indicates that the XGBoost based method is less affected by the lack of information about the later growing stages, possibly hinting that the underlying relations between early growing stages and final yield are better understood than by the Deep Learning methods. Regarding the hyperparameters of XGBoost, similar values are observed as described for the end-of-year prediction.

\begin{table}
\begin{tabularx}{\textwidth}{ |c| *{6}{Y|} }
\cline{2-7}
   \multicolumn{1}{c|}{} 
 & \multicolumn{2}{c|}{XGBoost}  
 & \multicolumn{2}{c|}{CNN}
 & \multicolumn{2}{c|}{CNN-LSTM}\\
\hline
Year & $\operatorname{RMSE}$ & $\operatorname{R^2}$ & $\operatorname{RMSE}$ & $\operatorname{R^2}$ & $\operatorname{RMSE}$ & $\operatorname{R^2}$\\
\hline
2017 & {\leavevmode\color[HTML]{009901}4.88}  & {\leavevmode\color[HTML]{009901}0.72} & {7.82}       & 0.28      & {7.78}         & 0.29         \\ 
2018 & {\leavevmode\color[HTML]{009901}5.27} & {\leavevmode\color[HTML]{009901}0.68} & 7.30 & 0.50 & 8.06 & 0.38 \\ 
2019 & {\leavevmode\color[HTML]{009901}4.65} & {\leavevmode\color[HTML]{009901}0.70} & 9.73 & -0.49 & 9.67 & -0.31 \\ 
2020 & {\leavevmode\color[HTML]{009901}5.35} & {\leavevmode\color[HTML]{009901}0.65} & 8.24 & 0.32 & 8.16 & 0.33 \\ 
2021 & {\leavevmode\color[HTML]{009901}6.04} & 0.61 & 8.11 & {\leavevmode\color[HTML]{009901}0.62} & 8.76 & 0.56 \\ 
AVG  & {\leavevmode\color[HTML]{009901}5.24} & {\leavevmode\color[HTML]{009901}0.67} & 8.24 & 0.25 & 8.49 & 0.25 \\
\hline
\end{tabularx}
\caption{\label{table:mis1} $\operatorname{RMSE}$ and $\operatorname{R^2}$ values of \textbf{\textit{in-year soybean yield predictions}}. The best result for every year is highlighted in green. The CNN architecture is taken from \citet{you2017deep}, and the CNN-LSTM architecture is taken from \citet{sun2019county}.}
\end{table}

\subsection{Data and Time Efficiency}
\label{sec:data_and_time_efficiency}
While the raw satellite images take up 218 GB, histogramization results in a reduction of down to 1.41 GB by removing spatial dependencies of the data. The histograms still cover the complete value ranges of the pixel intensities. However, since adjacent image pixels have a high probability of showing similar values, multiple bins of these histograms show zero entries. Our approach of estimating only three key values (i.e. median, 20\% quantile, and 80\% quantile) describing the value distributions produces a representation size of 0.01 GB for further processing.  
\par
The USDA yield dataset shows the most significant amount of training data that need to be processed for the year 2021. To examine the time used to train and test a yield prediction model for 2021, the performance is averaged over two runs, since the early stopping mechanic within the training of the Deep Learning approaches is responsible for varying training times. The CNN network requires approximately 15 minutes for training. The CNN-LSTM architecture takes approximately 10 minutes for the task. Finally, the XGBoost-based approach takes about 90 seconds for the same task. All experiments were performed on an NVIDIA GeForce GTX 1660 Ti. 

\subsection{Explaining Predictions with Feature Importances}
\label{sec:explainability}

A prediction model that users can trust must be equipped with the ability of explanation. This demand contrasts with the concept of ``black boxes'' in Machine Learning, where even the developers are often not able to explain the model's behavior. In particular, explainability integrated for Deep Learning architectures is still in its early stages. 

The term \textit{feature importance} refers to an importance score for all input features for a given prediction model. Input features with higher importances will have larger effects on the prediction results. Feature importances can support the credibility of prediction results by comparing them with established expert knowledge. Additionally, feature importances can help to save costs by identifying and dropping such features that, on the one hand, show low importance but are, on the other hand, expensive or time-consuming to obtain. There are several examples of feature importances, such as the permutation importance or Mean Decrease Accuracy (MDA) \citep{breiman2001random} or the Mean Decrease in Impurity (MDI) \citep{louppe2013understanding}. Recent research shows that a game theoretic approach to feature importances possesses desirable properties. The Shapley Value feature importances (SHAP values) compute optimal explanations that directly measure local feature interaction effects to understand the global model structure based on the combination of many local explanations of each prediction \citep{lundberg2020local}. For this evaluation, the Tree SHAP algorithm is used, which allows for a fast computation of local SHAP importances for tree structures.

\begin{figure}[ht]
	\centering 
	\includegraphics[width=\textwidth]{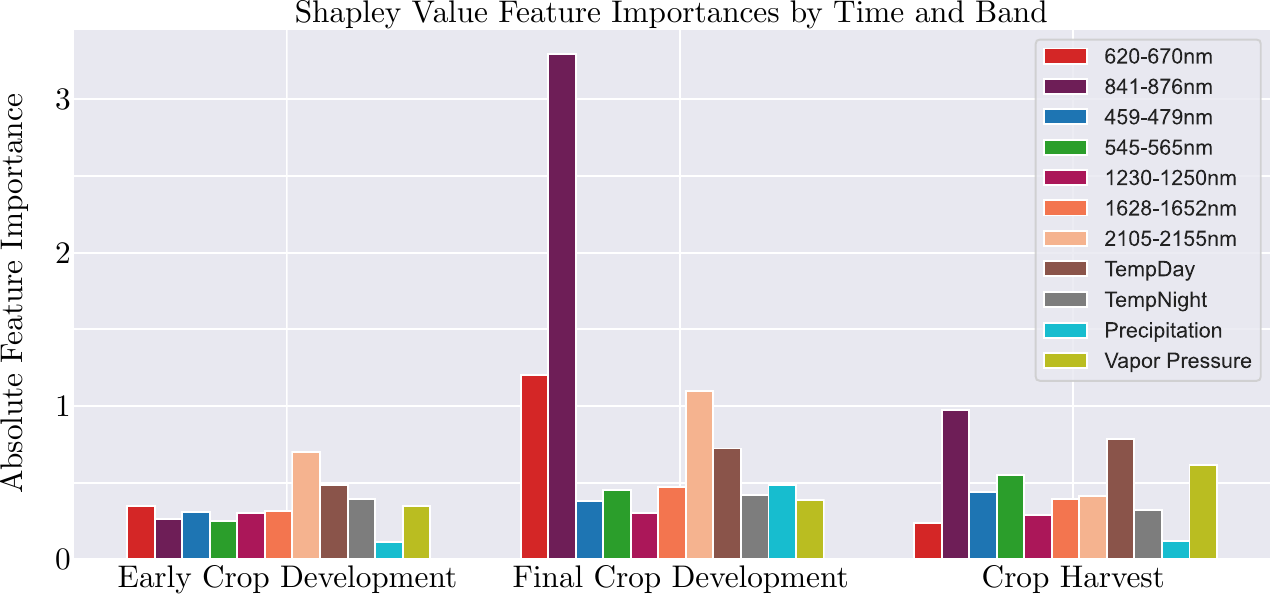}
	\caption{Shapley Value feature importances computed for the XGBoost-based prediction model used to predict the yields at the end of the year 2021. Importances are aggregated over multiple features to show the impact of the individual bands for three different timeframes. Namely, Early Crop Development from day 82 to day 161, Final Crop Development from day 162 to day 241 and Crop Harvest from day 242 to day 321.}
	\label{fig:importances}
\end{figure}

The local feature importances are calculated for all instances of the training dataset for the end-of-year yield prediction for 2021 and are shown in Figure \ref{fig:importances}. The feature importances are derived by averaging the absolute SHAP values over the complete training dataset. As described above, each data instance consists of 1122 features. Features that are closely related to each other are aggregated to obtain a meaningful representation of the feature importances. For each feature, it can be determined which of the initial 11 bands in which time step was used for its calculation. This information is utilized to group the feature importances according to the 11 bands and 3 general time periods each. Each time period spans 80 days, showing the influences of the features beginning on the 82nd day of the year. Since planting and harvest dates vary between the different years and states, broad descriptions of the state of the plant in each time frame are given. The first time period for early crop development spans the entire planting process and contains the earliest stages of growth. The second time period ends shortly before the first harvesting for most states and therefore shows the final crop development. The last time frame spans the harvest season, and at the end of it, all harvests are completed \citep{USDA_dates}.

Overall, the XGBoost-based prediction model shows high feature importances for the last and especially for the second selected time frame. This shows that information captured closer to harvest is the most important for feature selection and explains why the end-of-the-year prediction comes with better accuracies than the in-year prediction, where this information is missing. For the first time frame, all importances are low except for the seventh surface reflectance band with wavelengths between 2105 and 2155 nm, which is usually utilized in atmospheric studies and not in traditional crop monitoring. In the second and third time frames, the second MODIS reflectance band (purple) is the most significant. These wavelengths between 841 and 876 nm in the near-infrared spectrum are also widely known by experts to be very important for predicting crop yields \citep{quarmby1993use}. In terms of temperature and climate data, the analysis shows that temperature is more important than precipitation and vapor pressure, especially the temperature during the day. 
\par
In addition to the importances shown in Figure \ref{fig:importances}, an additional analysis of the impact of the handcrafted features defined in Section \ref{sec:hists_and_params} is conducted, showing that the average yield of the last five years is an essential feature for our model. This feature is able to give some spatial context to the data and helps to model geological differences.

\subsection{Discussion}
The experiments above hint towards a good compatibility between \mbox{XGBoost} and yield estimation, showing at least on par performance compared to Deep Learning for the end-of-year as well as for the in-year predictions. However, it is generally known that Deep Learning methods benefit from extensive training data. This tendency is especially noticeable on the CNN approach after expanding its training dataset following early experiments. While the XGBoost method showed similar results for the reduced and the full datasets, the accuracy of the CNN improved. Another indicator of this hypothesis is the good performance of the CNN approach for yield prediction in 2021, where most data are available. At the same time, 2021 is the only year in which the CNN could outperform XGBoost in terms of $\operatorname{R^2}$.

While having this in mind, the limitations of real world data need to be acknowledged. Arguably, soybean production in the United States is one of the largest datasets available to train and test models for yield prediction. Therefore, using the same approaches for other datasets will often result in having fewer data available to train the models. As more years pass, it will be possible to train the models on even larger datasets, with more harvest data available. Also, other possibilities for handling scarce data in Deep Learning could bring improvements. For classic image processing tasks, pretrained networks are utilized. The networks are pretrained to extract general features on big datasets in a task agnostic way. For a specific use case, the pretrained networks are trained further on often small task dependent use cases, achieving good accuracies on scarce datasets \citep{kolesnikov2020big}. Other research fields include semi-supervised learning, where only parts of the data are annotated \citep{pham2021meta}, and synthetic datasets, where more training data are generated \citep{xu2019modeling}.

Another point worth highlighting in this study is the application of different processing methods for the same data. Experiments have shown that the respective way of processing data – meaning histograms for Deep Learning and features for XGBoost – obtains the highest accuracies, respectively. The difference in accuracies between the CNN and the CNN-LSTM approaches must be explained. \cite{sun2019county} claim that the CNN-LSTM approach outperforms a CNN in their article. However, the CNN used for comparison consists only of the CNN part of the CNN-LSTM and is not geared specifically towards the problem. Reproducing their experiments showed the same results for the same CNN, but the CNN developed by \cite{you2017deep} is still able to outperform the CNN-LSTM on our data. 

Furthermore, the use of vegetation indices could bring further improvements in accuracy. Multiple studies are conducted in which remote sensing data are accessed for yield prediction, based on the results of different vegetation indices. Related work that uses remote sensing data through vegetation indices is described in Section \ref{sec:related}, both for Deep Learning and for classic Machine Learning applications. Within this study, it is expected that the information integrated within the vegetation indices is incorporated in the models directly. This is due to the assumed capabilities of all the approaches involved in our study to understand the relationships between raw remote sensing data and vegetation indices. The feature importances shown in Figure \ref{fig:importances} highlight that the XGBoost model makes use of the red and near-infrared bands of our data. Since these are the two bands used to calculate the NDVI it can be concluded that the model already uses the information that could be added by including vegetation indices like the NDVI. The feature importance analysis of \cite{you2017deep} shows similar results, hinting that the same conclusion can be drawn for the Deep Learning approaches. In addition to the NDVI, the Leaf Area Index (LAI) is a popular choice for yield prediction. The LAI measures the one-sided green leaf area per unit ground surface area and cannot be calculated as the aggregation of different surface reflectance bands directly. Estimating the LAI from remote sensing data is a difficult task in itself \citep{zhengLAI}. Since a model can be trained to infer the LAI from remote sensing data, it can be assumed, that the models within this study can comprehend the important information within the LAI for yield prediction directly. However, including several vegetation indices, such as NDVI or LAI, in the experiments offers an interesting research question for the future.

This study's results need to be interpreted with caution, due to some natural limitations of the used approach. All of the experiments are performed on a single dataset only, proving the capabilities of XGBoost for soybean yield prediction in the United States, in comparison to state-of-the-art Deep Learning approaches. However, this is only one possible task for predicting yields out of many. Most tasks are more difficult when considering varying amounts of training data and smaller study areas. The selection of crops within the United States allows the use of the CDL and Daymet data, which are not available worldwide. Although substitutable data sources are available around the world, the quality of the data and the resulting prediction accuracy may differ. Not only the region, but also the crop of interest could be different for further yield prediction scenarios. Further experiments will be needed to test the transferability of the insights within the study. The results are expected to be similar for a variety of crops, such as, for example, corn and wheat as they are planted and harvested in a yearly cycle and the production is set to maximize the yield. More difficult could be the prediction of yields from plants that are not planted yearly and therefore are bound to show interannual effects that the model does not need to account for while predicting soybean yields. This includes, for example, the prediction of grapes in viticulture or anything that grows on trees such as apples or citrus fruits. Furthermore, there are restrictions of using only remote sensing images as input, as other sources of information that could be beneficial for yield prediction, such as irrigation schedules, fertilization, or soil properties are left out. Finally, the limitations of XGBoost need to be acknowledged. Although very good prediction accuracies have been achieved in many real-world scenarios, some weaknesses are well known. This includes a lack of extrapolation capabilities when dealing with test data that exceed the observed feature quantities during training. Especially in times of climate change, this could pose future problems for our approach. Furthermore, XGBoost can be heavily affected by outliers due to the nature of Gradient Boosting, where each learner tries to consider the previous learners' mistakes. Since data acquisition of harvested yields always includes human-made measurements, mistakes and therefore outliers could occur. Given these points, future work includes deploying this study's approach on different datasets to validate XGBoost for yield prediction under different circumstances.


\section{Conclusions}
This study proposes a processing pipeline to efficiently process remote sensing images to feature vectors for Machine Learning. The experiments hint that the derived feature-based representation of the remote sensing data allows for the successful application of XGBoost for yield prediction. This is demonstrated by outperforming state-of-the-art Deep Learning based yield predictors on the USDA soybean dataset on multiple metrics and test cases. However, it also needs to be acknowledged that the possibilities for Deep Learning on yield prediction are not fully explored, and improvements are to be expected with more training data becoming available or with applying popular strategies for handling scarce data.

The trained models can be used to gain insight into the specific conditions in different application areas via feature importance analysis that is natively available due to the use of decision trees as a base learner for the XGBoost approach.

The approach to automated yield prediction based on efficient feature extraction from remote sensing data and XGBoost shows that:
\begin{itemize}
\item XGBoost is capable of yield prediction on remote sensing data on par with state-of-the-art Deep Learning approaches,
\item Shapley Value feature importances can raise trust in yield prediction models by exposing important features to be in line with experts knowledge on yield prediction.
\end{itemize}
\par
Although this study focused on soybean yield prediction in the United States, it is expected that this approach generalizes to a much broader range of yield prediction tasks. We are currently working to apply this approach to the prediction of grapevine yield. Additionally, the approach offers the opportunity to adapt to other prediction tasks in agriculture or other application areas.

\section*{\begin{normalsize}Acknowledgements\end{normalsize}}
\noindent
This work was partially done within the project ``Artificial Intelligence for innovative Yield Prediction of Grapevine” (KI-iRepro). The project is supported by funds of the Federal Ministry of Food and Agriculture (BMEL) based on a decision of the Parliament of the Federal Republic of Germany. The Federal Office for Agriculture and Food (BLE) provides coordinating support for artificial intelligence (AI) in agriculture as funding organisation, grant number FKZ 28DK128B20. 

We thank Timm Haucke and Frank Schindler for the proofreading of our manuscript. And Timm Haucke for fruitful discussions.

\bibliographystyle{elsarticle-harv} 
\bibliography{cas-refs}

\end{document}